\documentclass{esannV2}
\usepackage[latin1]{inputenc}
\usepackage{amssymb,amsmath,array}
\usepackage{graphicx} 
\usepackage{gensymb}
\begin{document}
\title{Deep Learning to Detect Bacterial Colonies\\for the Production of Vaccines}

\author{Thomas Beznik$^1$\thanks{T.B. was affiliated to UCL at the time of the work; he works now for the company RELU.}, Paul Smyth$^2$, Gael de Lannoy$^2$ and John A.~Lee$^1$\thanks{J.A.L.~is a Senior Research Associate with the Belgian F.R.S.-FNRS.}
\vspace{.3cm}\\
1- Universit\'e catholique de Louvain, Louvain-La-Neuve, Belgium
\vspace{.1cm}\\
2- GSK Vaccines, Rixensart, Belgium\\
}

\maketitle

\begin{abstract}
During the development of vaccines, bacterial colony forming units (CFUs) are counted in order to quantify the yield in the fermentation process. This manual task is time-consuming and error-prone. In this work we test multiple segmentation algorithms based on the U-Net CNN architecture and show that these offer robust, automated CFU counting. We show that the multiclass generalisation with a bespoke loss function allows distinguishing virulent and avirulent colonies with acceptable accuracy. While many possibilities are left to explore, our results show the potential of deep learning for separating and classifying bacterial colonies. 
\end{abstract}

\section{Introduction}
Vaccines are considered one of the greatest public health achievements and each year prevent 2 to 3 million deaths, with 86\% of the children worldwide getting vaccinated \cite{WHO_vaccines}. One important step in vaccine development is the measure of the number of viable bacterial colonies and the proportion of virulent colonies among them, as a vaccine requires a certain proportion of virulent bacteria to be effective. Samples are taken and applied onto Petri dishes containing a blood agar substrate and after a couple of days the single bacteria have grown into heaps of billions of cells called Colony Forming Units (CFUs) which are visible to the naked eye. The proportion of viable bacteria in the solution can thus be estimated by counting the number of CFUs, since each CFU originated from a single bacterium. The CFUs can be divided into two classes: virulent (also called bvg+) and avirulent (also called bvg-). These two classes can be distinguished through visual inspection: virulent colonies lyse the blood cells, i.e. destroy their membrane, which produces a dark halo around the colony, while avirulent colonies do not and thus no halo is observed. 

The purpose of this work is to automate the bacterial colony detection and identification process in order to free time for biologists to concentrate on other tasks, specifically to design a robust automated CFU counting algorithm that is able to distinguish bvg+ and bvg- colonies. 

\section{State of the art}
Automation of colony counting has been on the minds of researchers as early as 1957 \cite{mansberg1957automatic} and is still being actively researched to this day \cite{torelli2018autocellseg, jagga2018image, coconut_2018}. The methods can be grouped in three categories. First, hardware solutions applied directly to the plates that count by sweeping the plate with a scanning spot and observing the light transmitted via a photo-tube (e.g. \cite{mansberg1957automatic,alexander1958automatic}). These solutions are not autonomous as parameters need to be manually set for each plate.

Second, software solutions applied to images of the plates using various thresholding methods (see e.g.~\cite{chiang2015automated, uppal2012computational, torelli2018autocellseg} for recent works). The performance of such approaches are often assessed on limited, already-seen data, and they typically struggle to handle artefacts in the agar, reflections, bubbles, variability of the shape and color of the colonies and especially to separate touching cells. 

Finally, modern machine learning methods for counting bacterial colonies in images, following the work of \cite{flaccavento2011learning}, where random forests are used to identify individual colonies. The results of this method appear reasonable, but have been evaluated on only 9 images. The first deep learning, convolutional neural network (CNN) approach to colony counting was published in 2015 \cite{CFU_counting_CNN}, significantly outperforming other methods. More recently, in 2018, a fully convolutional approach was used \cite{andreini2018deep}.  Here the CNN was applied directly to the images, rather than to segments of the image. However, their approach did not count the number of colonies, but only separated background from colonies, thus avoiding problems related to touching colonies.  

In summary, there are several gaps in the literature that this work aims to address: first, to our knowledge, no study has been made on the classification of the colonies. Second, studies on deep learning solutions to CFU counting use old network architectures. Finally, there is no research on the use of a deep learning end-to-end solution, i.e. a network that receives an image of a Petri dish and outputs the colony counts and classes. 

\section{Data collection, experiments, and results}

\begin{figure}[ht!]
\centering
    \includegraphics[width=0.8\textwidth]{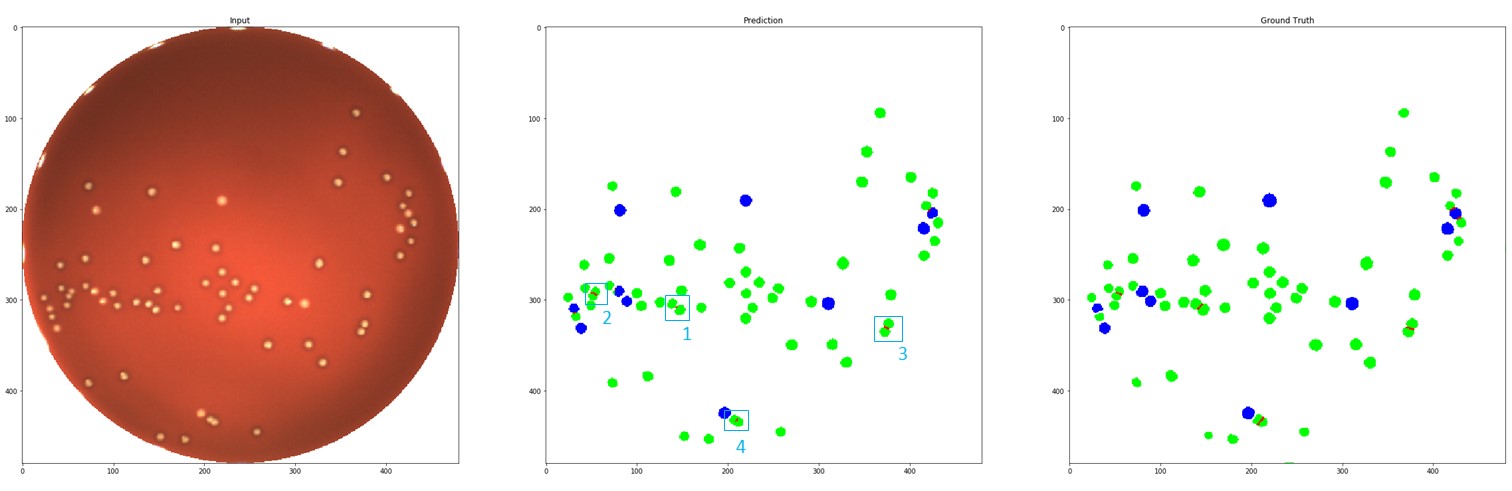}
    \caption{Left: Test image. Middle: Model prediction. Right: ground truth.}
    \label{fig:good}
\end{figure}

A total of 108 images of Petri dishes containing bacterial colonies were obtained from GSK laboratories using the laboratory aCOLyte camera (Synbiosis, Cambridge, UK). The plates are placed at a defined location and are illuminated by LEDs above and below the plate, and an image of size 480x480x3 in taken (see Fig. \ref{fig:good}, left). There are on average 4.726 `bvg-' colonies and 20.357 `bvg+' colonies in a single Petri dish. The dataset was labelled independently by five expert annotators. We define four classes that the pixels can belong to: background, bvg+, bvg- and border. The first three are the classes of interest, the border class has been introduced to handle the separation of touching colonies.

The first difficulty is the dataset size: there are only 108 images, and the models will train on even fewer images due to cross-validation.
We chose to work with such a small dataset in order to investigate if deep learning could be applied on this problem without requiring extensive data labelling.
Most deep learning architectures for computer vision involve millions of parameters and use very large datasets to train. Using these architectures will likely result in overfitting: we thus need to find architectures that are well fitted for limited datasets, use transfer learning, data augmentations, and other techniques.

A second difficulty is the imbalance of the classes in the dataset: 98.927\% of the pixels are of the class background, 0.825\% are bvg+, 0.238\% are bvg- and 0.009\% are border. We can thus see that there is a very high class imbalance between the background pixels and the other classes. This is due to the fact that the colonies are small and in small numbers, while the rest of the image is considered as background. A trivial model that predicts background for every pixel would obtain a training accuracy of 98.927\% without detecting any colony. We must thus take this class imbalance into account in the training and validation phase to avoid this issue; this will mostly be handled by adapting the loss.

As seen in Fig.~\ref{fig:good}, a third challenge is the presence of noise in the images, mainly the reflections of the LEDs on the border of the agar plate: these reflections could be easily mistaken for colonies and many watershed-based algorithms would struggle to discard them. We must also address the challenge of separating touching colonies, which can be complicated even for humans.


The dataset is randomly split into training-validation (80\%) and test (20\%). 
Random rotations in 
$[0\degree , 360\degree ]$ were used to augment the small amount of training data. 
Two network architectures are explored in the hyperparameter search: 

\begin{enumerate}
    \item Regular U-Net \cite{UNet}: the hyperparameters that are searched are the number of layers (2, 4 and 6) and the use of batch normalization.
    \item U-Net with pre-trained encoder \cite{ImageNet}: the only hyperparameter is the pre-trained network (ResNet-50, ResNet-152, Inception-ResNet-v2, or DenseNet-169). The number corresponds to the number of layers of the network.
\end{enumerate}

Two losses are investigated in the hyperparameter search: the weighted cross-entropy and the categorical cross-entropy with soft DICE per channel \cite{kodym2018segmentation}. Both are composed of weights that constitute additional hyperparameters. These losses are adapted to the inherent class imbalance. For all models, the Adam optimizer is used and several learning rates are studied (1e-3, 1e-4, and 1e-5). A mini-batch size of 8 is used for the regular U-Net models and of 4 for the pre-trained U-Net models. Early stopping is used with a patience of 10 epochs.

The results were collected in two phases: hyperparameter search and best model evaluation. In the first phase, the impact of all the hyperparameters defined above is assessed through a 4-fold cross-validation. Their performance is evaluated using the ``mAP (mean Average Precision) over multiple IoU thresholds" metric \cite{Coco_challenge}. This metric defines a true positive as an object which is detected with an IoU above a certain threshold. The metric is then computed by averaging the Average Precision over multiple IoU thresholds, typically from 0.5 to 0.95 with a step of 0.05.
For the second phase, the hyperparameter set that obtained the best validation ``mAP over multiple IoU thresholds" is selected, it is trained on the full training-validation set and evaluated on the test set. We use the ``MAE (Mean Average Error)" metric to judge the colony counting accuracy of the best model. The colony count is obtained by counting the number of connected components of each class in the mask, ignoring the `border' pixels.

The most influent hyperparameters were the learning rate and the type of architecture. The best performing architecture is the ResNet-152 pre-trained U-Net encoder, trained with a learning rate of 1e-4 and a weighted cross-entropy loss with weights $w_\mathrm{background}=0.01$, $w_\mathrm{bvg+}=0.25$, $w_\mathrm{bvg-}=0.34$ and $w_\mathrm{border}=0.4$. General statistics of the best model are shown in Table \ref{train_stats}. We can see that the performances on the `border' class are worse than those for the `bvg+' and `bvg-' classes. For the training set, the precision on this class is 0.758 and the recall is equal to 0.518, which is already not satisfactory. The model obtains a `border' precision of 0.503 and recall of 0.285 on the test set. The model clearly has not learned well to detect border pixels and is thus not good at separating touching colonies (see Fig \ref{fig:good}). We can also observe in Table \ref{train_stats} that the precisions and recalls for the `bvg+' and `bvg-' pixels are high, both on the training and test sets.
Finally, we can see that the MAE, the average difference in colony counting per image, roughly doubles from the training set to the test set for the `bvg+' and `bvg-' colonies: for the test set, the model is on average off by 1.048 `bvg+' colony and 0.143 `bvg-' per image. We can also see that the `bvg-' MAE is much smaller than the `bvg+' MAE for the same dataset; the model thus seems to make fewer counting mistakes for the `bvg-' colonies than for the `bvg+' for one image, but we believe that this is biased by the fact that there are generally much more `bvg+' colonies than `bvg-' in an image.

\begin{table}[h]
\centering
\begin{tabular}{c|c|c|cc}
Class & Dataset & MAE   & Precision & Recall \\
\hline
Bgv+ & Training & 0.798 & 0.952     & 0.961 \\
Bgv+ & Test     & 1.048 & 0.906     & 0.927 \\
Bgv- & Training & 0.059 & 0.955      & 0.961 \\
Bgv- & Test     & 0.143 & 0.919     & 0.905\\
Border & Training & & 0.758          & 0.518\\
Border & Test & & 0.503          & 0.285\\
\end{tabular}
\caption{Statistics of the best model on the training and test sets.}
\label{train_stats}
\end{table}

\begin{figure}[ht!]
\centering
    \includegraphics[width=0.85\textwidth]{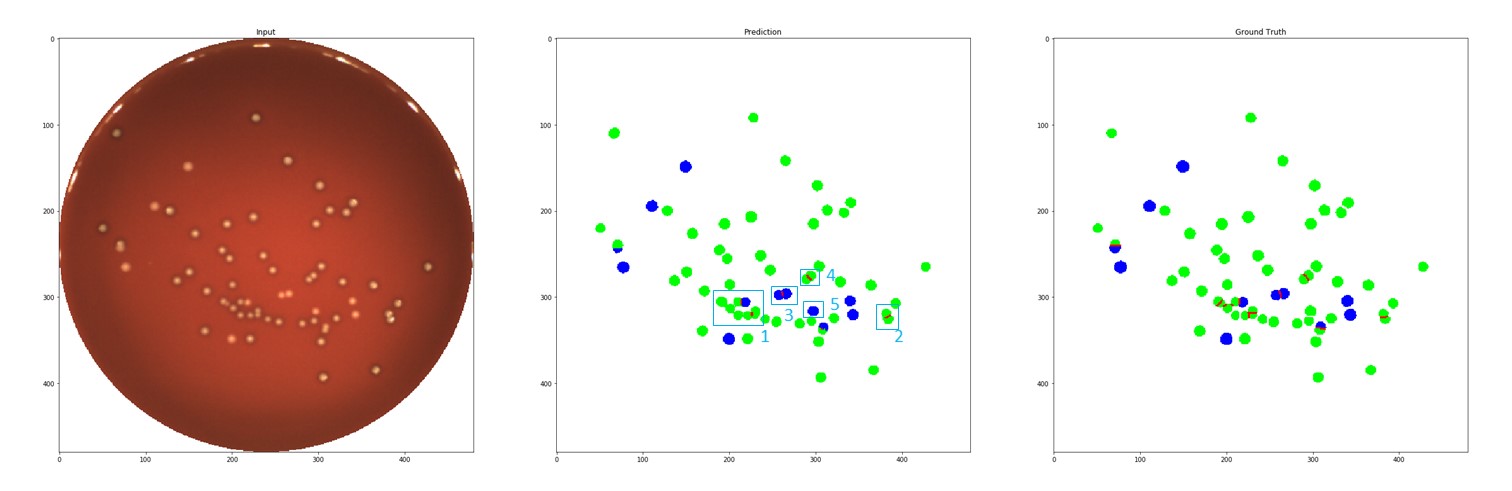}
    \caption{Left: Test image. Middle: Model prediction. Right: ground truth.}
    \label{fig:bad}
\end{figure}

Figures \ref{fig:good} and \ref{fig:bad} show two outputs of the model on images of the test set. These outputs illustrate the strengths and weaknesses of the model on unseen data. We identify the colonies of interest in the images using blue boxes and an associated number. First, we see that for both images the model detects and classifies the colonies quite well; there is only one class error in Fig.~\ref{fig:bad}, for colony number 5. This is actually a labelling error; the colony is indeed `bvg-', as the model predicted. We can also observe how the model handles the separation of colonies for these input images. In Fig.~\ref{fig:good}, we see that colonies 1, 2, and 3 are well separated with `border' pixels, but that the colonies at 4 are not completely separated. They will thus be counted as one colony. In Fig.~\ref{fig:bad}, the separation of the colonies at 1 is badly handled by the model and there will thus be a high counting error for this image. On the other hand, colonies 2, 3, and 4 are well separated. The incomplete separation of touching colonies is the most common error of the model. The model often detects some `border' pixels between touching colonies, but it fails to completely separate them, which results in a counting error. Simple post-processing could greatly improve the separation by for example extending the border. Finally, we see that for both images the model ignores the reflections on the border of the Petri dish. In general, the model is good at detecting, classifying and identifying the correct area of the colonies.
 
\section{Conclusion and future works}
We have shown that deep learning algorithms can overcome the drawbacks of existing computer vision solutions for classification of bacterial colonies in images of Petri dishes with access to a limited dataset. The architecture U-Net \cite{UNet} with a pre-trained ResNet-152 \cite{resnet} encoder, weighted cross-entropy loss, combined with image augmentations, achieves promising results on unseen images. We have shown that this approach can overcome some of the drawbacks of existing computer vision solutions: it is robust to reflections, can consistently detect colonies of various shapes and sizes and it achieves complete automation. 
The proposed approach contributes to closing a gap in the literature: no study, to our knowledge, has explored the use of deep learning to solve the problem of CFU counting end to end, including the classification of bacterial colonies. These findings open the way to a new approach of to automating bacterial colony counting and classification, fitting into the bigger movement towards full laboratory automation. Future research will investigate an improved separation of touching colonies by using specific losses such as the focal loss \cite{lin2018focal}, ``U-Net" loss \cite{UNet}, or changing the task to bounding-box detection. It will also be interesting to evaluate a larger and more diverse dataset, perform a broader hyperparameter search and to study the benefit of post-processing methods.  

\section{Disclosure}
This work was sponsored by GlaxoSmithKline Biologicals SA. All authors were involved in drafting the manuscript and approved the final version. The authors declare the following interest: GDL and PS are employees of the GSK group of companies and report ownership of GSK shares and/or restricted GSK shares.

\begin{footnotesize}

\bibliographystyle{unsrt}
\bibliography{bibliography}

\end{footnotesize}


\end{document}